\documentclass[times, 10pt,twocolumn]{article} 
\usepackage{latex8}
\usepackage{times}
\usepackage{epsfig}
\usepackage{amsmath}
\usepackage{mathtools}
\usepackage{units}
\usepackage{url}
\usepackage{titling}
\usepackage{titlesec}
\newcommand{\specialcell}[2][c]{%
  \begin{tabular}[#1]{@{}c@{}}#2\end{tabular}}

\begin{document}

\title{Machine Learning on Biomedical Images: Interactive Learning, Transfer Learning, Class Imbalance, and Beyond}

\author{Naimul Mefraz Khan\qquad Nabila Abraham \qquad Marcia Hon \qquad Ling Guan \\Ryerson Multimedia Research Laboratory \\ Ryerson University, Toronto, ON \\ E-mail: n77khan@ryerson.ca}
\maketitle
\thispagestyle{empty}

\begin{abstract}
 In this paper, we highlight three issues that limit performance of machine learning on biomedical images, and tackle them through 3 case studies: 1) Interactive Machine Learning (IML): we show how IML can drastically improve exploration time and quality of direct volume rendering.   2) transfer learning: we show how transfer learning along with intelligent pre-processing can result in better Alzheimer's diagnosis using a much smaller training set 3) data imbalance: we show how our novel focal Tversky loss function can provide better segmentation results taking into account the imbalanced nature of segmentation datasets. The case studies are accompanied by in-depth analytical discussion of results with possible future directions.  
	\end{abstract}

\Section{Introduction}

Machine Learning (ML) has shown promising advances in the field of biomedical imaging, ranging from applications such as computer-aided diagnosis and image segmentation. Recently, deep learning techniques such as Convolutional Neural Networks (CNN) have provided significant performance boost to the aforementioned fields, at times beating the performance of human experts. Despite such impressive performance, the adoption of ML in clinical practice has been slow. Over the last few years, we have identified the following key issues preventing widespread adoption of ML for analysis of biomedical images:\newline
1. \textit{Interactivity}: ML systems are typically designed as a black box, where the end-user does not have any means of interpreting the results easily, or interacting with the system. This is a major issue that especially prevents adoption of ML systems in clinical practice. For clinical application the system should be transparent and interactive \cite{kwon2019retainvis}. \newline
2. \textit{Dependence on large training dataset}: deep learning model requires a large number of training samples, which can be a problem for the biomedical imaging field where physician-annotated data can be hard to acquire. They also require massive parameter tuning and huge amount of computational resources. \newline
3. \textit{Class imbalance}: Class imbalance is also a common scenario in biomedical imaging, such as having smaller number of positive cases for computer-aided diagnosis, or smaller number of lesion pixels for segmentation \cite{focalloss}. This imbalanced nature of data can result in reduced performance for ML algorithms.

In this paper, we summarize three case studies that we have undertaken in the past few years to tackle these issues. While the solutions we propose are not fool-proof, they pave the way towards more robust ML algorithms and systems that are ready for deployment in clinical practice. 

In Case Study I, we propose a novel direct volume rendering system \cite{rmlneucom} that employs the philosophy of Interactive Machine Learning (IML) \cite{iml}, where the target is to design a system that will allow the end-user to directly interact with the ML system through a well thought-out interactive UI. In the proposed system, we utilize a novel Self-Organinzing Map (SOM) architecture \cite{som} with an intuitive UI which enables quick iteration through direct volume rendered 3D images. We show that our proposed system drastically cuts down required time to generate a satisfactory visualization. 

In Case Study II, we address the issue of large training dataset through transfer learning \cite{transfer, rmlbibm}, where we utilize a state-of-the-art CNN architecture pre-trained on natural images and show that through intelligent training data selection based on entropy criterion, this network can be fine-tuned with only a small training dataset for computer-aided diagnosis. We present results on the state-of-the-art ADNI dataset \cite{adni} and show that transfer learning provides significantly improved results for Alzheimer's diagnosis. 

In Case Study III, we show how the issue of class imbalance can be dealt with using a novel loss function \cite{rmlisbi}, where we achieve a balance between precision and recall using the Tversky index \cite{tversky} combined with the focal loss function \cite{focalloss}, achieving state-of-the-art results in breast ultrasound segmentation.

Finally, we provide some concluding remarks on how to further refine the proposed methods for a holistic approach towards ML for biomedical imaging.      

\Section{Case Study I: Interactive Machine Learning }

The target of Interactive Machine Learning (IML) is to involve the end-user in the ML process \cite{iml}. Through utilizing innovative human-computer interaction techniques, IML enables the end-user to control parameters and behaviour of ML algorithms easily and efficiently. IML can be particularly useful in healthcare, since end-user (healthcare practitioners) involvement in the final decision making process is crucial \cite{kwon2019retainvis}. In this case study (originally published in \cite{rmlneucom}), we show that IML can drastically simplify a task that is frequently encountered in biomedical image visualization: Direct Volume Rendering (DVR). DVR makes use of a Transfer Function (TF), which maps one or more features extracted from the data (the feature space) to different optical properties such as color and opacity. DVR has become a popular technique for MRI and CT visualization.

The TF design is typically a user-controlled process, where the user interacts with different widgets (usually representing feature clusters or 1D/2D histograms) to set color and opacity properties \cite{multi2d,multiclustering,multisemi}. However, interacting with the feature space is difficult for the end-user, who may not have any knowledge about feature extraction and clustering. Multi-dimensional feature spaces can not represent distinguishable properties such as peaks and valleys which are important for proper TF generation from histogram \cite{multinonparam}. As a result, the user requires extensive knowledge about the actual volume and the TF widgets themselves. Also, these kind of widgets try to represent the feature space directly, putting a strict restriction on the type of features used and the dimensionality. There is a need for a more robust method so that any feature can be represented to the user in a visual form while maintaining the topological relationship between various data distributions. Hence, a good DVR system should be: 1) intuitive and easy to use without extensive technical experience, 2) time efficient by providing immediate results, 3) have separate classification and feature modules independent of each other so that any type of feature can be integrated easily and efficiently and 4) visually pleasing.

An ML method that can satisfy all the aforementioned requirements is Self-Organizing Map (SOM) \cite{som}. SOM is an unsupervised clustering algorithm that maps continuous input feature space patterns to a set of discrete output nodes. The nodes are arranged in a pre-defined lattice. Each node has a weight vector associated with it. The input patterns (or feature vectors) are assigned to the closest node in the Euclidean sense. In other words, the nodes ``compete'' among themselves to be representative of the underlying input patterns. As a result, at the end of the training cycle, the node lattice represents a topology-preserving low-dimensional clustered representation of the original feature space. They can also be colored in a very intuitive way by using U-matrices \cite{som}, so that the end-user can quickly identify underlying clusters. Since SOM nodes have direct correspondence with the original input data (voxels in this case), an end-user can select the clusters he or she deems to be important. In this way, the user will have full control over the regions to be defined as important. Due to the ability to use higher-dimensional features, the cluster representation can be sophisticated so that user interest is reflected in the grouping of voxels.

However, mere correspondence between SOM nodes and voxels is not enough for effective volume rendering. We need to generate the color and opacity properties to be assigned to the different clusters selected by the user. The typical trend is to let the user control the color and opacity specifications \cite{multiroettger}. Especially the color selection is mostly left out as a user task. However, this requires some visualization knowledge which the end user may not possess. In our proposed method, we calculate the opacity and color values automatically from the user selection of voxel groups. The opacity value is generated based on the variance of a group, since variance is related to visibility \cite{quick2ins}. For the color values, we take help of \textit{harmonic colors}. Harmonic colors are a sets of colors, the internal relationship among which provides a pleasant visual perception \cite{quick2ins}. We also assign the saturation and brightness of each voxel based on its properties \cite{rmlneucom}. In this way, by simply interacting with the SOM lattice, a user can generate the desired DVR quickly and efficiently.

We use the Spherical SOM structure (SSOM) \cite{sphericalsomthesis} because of it's restriction-free structure. The SSOM learns unsupervised clustering of voxel features (3D voxel coordinates, intensity value, 3D gradient magnitude, and second derivative). After the SSOM training, the user is presented with a visual form of the spherical lattice.The U-Matrix based coloring of the lattice represents existence of potential clusters, but we leave the final decision of grouping the nodes and defining them as a single group to the user. The user achieves this by dragging a rubber-band control on top of the visualized SSOM lattice. When the user is satisfied with the selection, he or she can define the current selection as one group, and do further selections for the next group. As soon as a group is defined, the rendering is updated to show the final result based on the properties of the voxels in the group and color harmonization.

\begin{figure}[htbp]
\centering
\epsfig{file=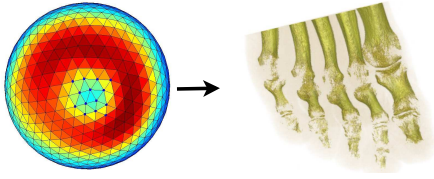,width=6cm}
\caption{Rendering of the bone structures from the Foot dataset}
\label{fig:som}
\end{figure}

Figure \ref{fig:som} shows a sample result, where the user successfully isolated the bone structures from the CT scan of a foot (\cite{vollibrary}). The cluster corresponding to the bone structure is clearly identifiable without any ML knowledge (the blue region in the middle representing density of unique voxels separated from other regions by the red area). With a traditional TF editor, this process would have taken much longer. 

To further validate our method, we performed user study with 5 users with different levels of computer expertise. The users were asked to generate a reference volume rendering with our proposed method and a traditional DVR editor. 
\begin{table}[htbp] {
\footnotesize
\centering
\begin{tabular}{|l|c|c|c|c|}
\hline
System & \specialcell{Training \\ Time} & \specialcell{Interaction \\ Time} & \specialcell{Interaction \\ Rating} & \specialcell{Output \\ Rating} \\ \hline
Proposed & \textbf{126.67} & \textbf{160} & \textbf{4.5} & \textbf{4.5} \\ \hline
 Traditional & 320 & 400 &3.33  & 2.5\\

\hline

\end{tabular}
\caption{Average Time (seconds) and rating (out of 5) for user training and interaction.}
\label{table:users}

}
\end{table}

Table \ref{table:users} shows the average times recorded and the ratings provided by the users for creating the reference rendering of the Foot dataset. The ``training time'' corresponds to average time required by the users to be comfortable with the systems, while ``interaction time'' corresponds to how long the users required on average to generate the rendering results. As we can see, our proposed method requires substantially shorter period of time for both training and interaction, while providing a richer user experience and output quality. Therefore, through this case study we show that IML has the potential to significantly improve tasks such as DVR by involving the end-user in the process. 

\Section{Case Study II: Transfer Learning}

In this Case Study, we show that by cleverly utilizing Transfer Learning (TL) combined with an intelligent training data selection process, we can reduce the dependency on a large training dataset that is typical for a Deep Learning (DL) method. The area of application we target is Alzheimer's Disease (AD) diagnosis. Early diagnosis of AD can be achieved through automated analysis of MRI images with ML. It has been shown recently that in some cases, ML algorithms can predict AD better than clinicians \cite{kloppel2008accuracy}, making it an important field of research for computer-aided diagnosis.

While statistical ML methods such as Support Vector Machine (SVM) \cite{plant2010automated} have shown early success in automated detection of AD, recently DL methods such as Convolutional Neural Networks (CNN) and sparse autoencoders have outperformed statistical methods. However, the existing DL methods train deep architectures from scratch, which has a few limitations \cite{finetunesurvey}: 1) properly training a DL network requires a huge amount of annotated training data, which can be a problem especially for the medical imaging field where physician-annotated data can be expensive to acquite; 2) training a deep network with large number of images require huge amount of computational resources; and 3) deep network training requires careful and tedious tuning of many parameters, sub-optimal tuning of which can result in overfitting/underfitting, and, in turns, result in poor performance.

An attractive alternative to training from scratch is fine-tuning a deep network (especially CNN) through Transfer Learning (TL) \cite{transfer}.  In popular computer vision domains such as object recognition, trained CNNs are carefully built using large-scale datasets such as ImageNet \cite{imagenet}. The idea of TL is to train an already-trained (pre-trained) CNN to learn new image representations using a smaller dataset from a different problem. It has been shown that CNNs are very good feature learners, and can generalize image features given a large training set. If we always train a network from scratch, this attractive property of CNN is not being utilized, especially given the popularity of CNN and the existence of proven architectures and datasets.

In this case study, we investigate how TL can be applied for improved diagnosis of AD. The key motivation behind employing TL is to reduce the dependency on a large training set. Due to the popularity of CNN, there are many established architectures that have been carefully constructed by researchers over the last few years to solve visual classification problems. The benchmark for evaluating the best architectures has been the ImageNet Large Scale Visual Recognition Challenge (ILSVRC), where the participants are given the task to classify images of 1000 different objects \cite{imagenet}. The thorough evaluation nature of the ILSVRC challenge ensures that the architectures that are ranked top in terms of performance are robust. We investigated the recent winners of the ILSVRC challenge to identify an architecture that will be suitable for AD diagnosis. We closely follow the VGG architecture \cite{vgg16} proposed by the Oxford Visual Geometry Group which won the ILSVRC 2014 challenge. The reason behind following the VGG architecture is not only the high accuracy, but also the efficiency, and more importantly, adaptability to other image classification problems than ImageNet \cite{vgg16}.  The convolutional layers of the architecture are used as feature extractors and kept fixed, and only the fully-connected layers are trained on the training data, which allows the method to learn robust feature variations with only a small training set.

Merely choosing training data at random may not provide us with a dataset representing enough structural variations in MRI. Instead, we pick the training data that would provide the most amount of information through image entropy.In general, for a set of $M$ symbols with probabilities $p_{1},p_{2},\ldots,p_{M}$ the entropy can be calculated as follows:
 \begin{align}
  \label{eq:entropy}
  H=-\sum_{i=1}^{M}p_{i}\log p_{i}.
  \end{align}
  
For an image (a single slice from an MRI scan), the entropy can be similarly calculated from the histogram \cite{entropy}. The entropy provides a measure of variation in a slice. The higher the entropy of an image, the more information it contains.  Hence, if we sort the slices in terms of entropy in descending order, the slices with the highest entropy values can be considered as the most informative images, and using these images for training will provide robustness. 

\begin{table}[htbp] {
\footnotesize
\centering
\begin{tabular}{|l|c|c|c|c|c|}
\hline
Method & \begin{tabular}{@{}l@{}}Training Size \\ (\# images) \end{tabular} & \specialcell{AD \\ vs. NC} & \specialcell{AD \\ vs. MCI} & \specialcell{MCI \\ vs. NC}  \\

\hline

\begin{tabular}{@{}l@{}}Patch-based \\ Autoencoder \cite{icml}\end{tabular} & 103,683 & 94.74 & 88.1  & 86.35 \\ 
\hline
3D CNN \cite{hosseini2016alzheimer} & 39,942 & 97.6 & 95 & 90.8\\ 
\hline
Inception \cite{deepad} & 46,751 & 98.84 & - & - \\
\hline
Proposed & \textbf{2,560} & \textbf{99.36} & \textbf{99.2} & \textbf{99.04} \\ 
\hline

\end{tabular}
\caption{Comparison of \% accuracy with contemporary methods on the ADNI dataset. }
\label{table:comp}
}
\end{table}
In Table \ref{table:comp}, we show that through intelligent training data selection and TL, we can achieve state-of-the-art classification results for all three classification scenarios in Alzheimer's prediction, namely, AD vs Normal Control (NC), Mild Cognitive Impairment (MCI) vs. AD, and MCI vs. NC on the benchmark ADNI dataset \cite{adni} \footnote{For further experimental details, see \url{https://github.com/marciahon29/AlzheimersProject/}}. The most significant contribution here is the training data size. We see that the training size for the proposed method is almost 16 times smaller than \cite{hosseini2016alzheimer}, the closest competitor. Utilizing TL and our intelligent data selection method, we can cut down on the dependency on large training sets, making deep learning methods more practical for clinical usage.
    
\Section{Case Study III: Class Imbalance}

Our final case study addresses the issue of class imbalance, which is a common problem in the biomedical imaging field. Class imbalance is especially prevalent for lesion segmentation, where the lesions typically occupy a very small fraction of the full image. Such imbalance in the data can lead to instability in established generative and discriminative frameworks \cite{sudre}. In recent literature, CNNs have been successfully applied to automatically segment 2D and 3D biological data \cite{sudre}. Most of the current deep learning methods derive from a fully convolutional network architecture (FCN), where the fully connected layers are replaced by convolutional layers. The popular U-Net is an FCN variant which has become the defacto standard for image segmentation due to its multi-scale skip connections and learnable up-convolution layers \cite{ronne}. 

Some recent works attempt at dealing with class imbalance for segmentation. The focal loss function proposed in \cite{focalloss} reshapes the cross-entropy loss function with a modulating exponent to down-weight errors assigned to well-classified examples. The focal loss prevents the vast number of easy negative examples from dominating the gradient to alleviate class imbalance. In practice however, it faces difficulty balancing precision and recall due to small regions-of-interest (ROI) found in medical images. Research efforts to address small ROI segmentation propose more discriminative models such as attention gated networks \cite{oktay}. 

To address the aforementioned issues, we combine attention gated U-Net with a novel variant of the focal loss function, better suited for small lesion segmentation. We propose a novel focal Tversky loss function for highly imbalanced data and small ROI segmentation, where we modulate the Tversky index \cite{tversky} to improve precision and recall balance; and combine it with a deeply supervised attention U-Net \cite{oktay}, improved with a multi-scaled input image pyramid for better intermediate feature representations. 

We extend from the Dice Loss (DL) function described in \cite{sudre}, whose limitation is that it equally weighs false positive (FP) and false negative (FN) detections. In practice, this results in segmentation maps with high precision but low recall. With highly imbalanced data and small ROIs such as lesions, FN detections need to be weighted higher than FPs to improve recall rate. The Tversky similarity index is a generalization of the Dice score which allows for flexibility in balancing FP and FNs:

\begin{align}
  TI_c = \frac{\sum_{i=1} ^N p_{ic} g_{ic} + \epsilon}{\sum_{i=1} ^N p_{ic}g_{ic} + \alpha\sum_{i=1} ^N p_{i\bar{c}}g_{ic} + \beta\sum_{i=1} ^N p_{ic}g_{i\bar{c}} + \epsilon} 
\label{eq:tversky} 
\end{align}

where, $p_{ic}$ is the probability that pixel $i$ is of the lesion class $c$ and $p_{i\bar{c}}$ is the probability pixel $i$ is of the non-lesion class, $\bar{c}$. The same is true for $g_{ic}$ and $g_{i\bar{c}}$, respectively. The total number of pixels in an image is denoted by $N$. Hyperparameters $\alpha$ and $\beta$ can be tuned to shift the emphasis to improve recall in the case of large class imbalance. The Tversky index is adapted to a loss function (TL) in \cite{tversky} by minimizing  $\sum_{c} 1 - TI_c$. 

Another issue with the DL is that it struggles to segment small ROIs as they do not contribute to the loss significantly. To address this, we propose the Focal Tversky Loss Function (FTL), parametrized by $\gamma$, for control between easy background and hard ROI training examples. In \cite{focalloss}, the focal parameter exponentiates the cross-entropy loss to focus on hard classes detected with lower probability. This idea has been extended in recent works where an exponent is applied to the Dice score \cite{wong}. Similarly, we define our FTL function as:

\begin{align}
 FTL_c = \sum_{c} (1 - TI_c)^{\nicefrac{1}{\gamma}}
\label{eq:FTL} 
\end{align}

where $\gamma$ varies in the range $\{1,3\}$. In practice, if a pixel is misclassified with a high Tversky index, the FTL is unaffected. However, if the Tversky index is small and the pixel is misclassified, the FTL will decrease significantly. 

When $\gamma>1$, the loss function focuses more on less accurate predictions that have been misclassified. However, we observe over-suppression of the FTL when the class accuracy is high, usually as the model is close to convergence. We experiment with high values of $\gamma$ and observe the best performance with $\gamma = \frac{4}{3}$ and therefore train all experiments with it. To combat the over-suppression of the loss function, we train intermediate layers with the FTL but supervise the last layer with the Tversky loss to provide a strong error signal and mitigate sub-optimal convergence. We hypothesize using a higher $\alpha$ in our generalized loss function will improve model convergence by shifting the focus to minimize FN predictions. Therefore, we train all models with $\alpha=0.7$ and $\beta=0.3$. It is important to note that in the case of $\alpha$ = $\beta$ = 0.5, the Tversky index simplifies to the DSC. Moreover, when $\gamma = 1$, the FTL simplifies to the TL.

To achieve further balance between precision and recall, we propose an improved attention U-Net \cite{oktay} that incorporates the proposed FTL. This architecture is based on the popular U-Net which has been designed to work well with very small number of training examples. At the deepest stage of encoding, the network has the richest possible feature representation. However, with cascaded convolutions and nonlinearities, spatial details tend to get lost in the high-level output maps. This makes it difficult to reduce false detections for small objects that show large shape variability \cite{oktay}. To address this issue, we use soft attention gates (AGs) to identify relevant spatial information from low-level feature maps and propagate it to the decoding stage. Moreover, since different kinds of class details are more easily accessible at different scales, we inject the encoder layers with an input image pyramid before each of the max-pooling layers. Combined with deep supervision, this method improves segmentation accuracy for datasets where small ROI features can get lost in cascading convolutions and facilitates the network learning more locality aware features with respect to the classification goal.

\begin{table}[htbp]

\centering 
\begin{tabular}{|l| c | c | c | c |} 

\hline  
Model & DSC  & Precision & Recall  \\ \hline 
U-Net + DL  & 0.547  & 0.653 & 0.658 \\ \hline
U-Net + FTL  & 0.669 & 0.775  & 0.715  \\\hline
Attn U-Net + DL & 0.615  & 0.675  & 0.658  \\\hline
\specialcell{Proposed (Attn U-Net + \\ Multi-Input + FTL)} &  \textbf{0.804}  & \textbf{0.829}  & \textbf{0.022} \\ \hline

\end{tabular}
\caption{Performance of the proposed FTL function on BUS 2017 Dataset B} 

\label{table:bus} 
\end{table}

Experiments were performed on the Breast Ultrasound Lesions 2017 dataset B (BUS) \cite{yap} \footnote{For further experimental details, see \url{https://github.com/nabsabraham/focal-tversky-unet}}, where the lesions occupy  4.84\% $\pm$ 5.43\%. As we see in Table \ref{table:bus}, when compared to the baseline U-Net, our methods improves Dice scores by 25.7\%, signifying the importance of addressing class imbalance for segmentation problems. 
 
\Section{Conclusion}
In this paper, we highlight three critical issues for ML in biomedical imaging, and summarize three case studies that attempts at solving these issues through utilization of several contemporary concepts. to involve the end-user in the ML process so that ML systems can be more trustworthy, we employ Interactive Machine Learning (IML), where the user can control the ML process through intuitive user interfaces. We present a case study on Direct Volume Rendering and show how IML can drastically improve the task. To combat the issue of dependence on large training sets for deep learning architectures, we employ Transfer Learning (TL) in combination with an intelligent training data selection method based on image entropy to fine-tune the state-of-the-art CNN architecture VGG for early diagnosis of Alzheimer's Disease. We show that the proposed method can achieve state-of-the-art performance with a training size almost 16 times smaller than the current methods. Finally, we address the class imbalance in medical image segmentation through a novel Focal Tversky Loss function, which can strike a good balance between precision and recall by incorporating the Tversky index into a focal loss function. We combine the proposed loss function with an attention U-Net architecture to achieve state-of-the-art results in breast ultrasound lesion segmentation on a benchmark dataset. 

The presented case studies only provide initial pathways into solving the issues. More work needs to be done to robustly solve these issues so that ML can be more suitable for deployment in clinical practice when dealing with automated analysis of biomedical images. For IML, we are looking into how the proposed solution can be adapted to deep learning methods. More importantly, we are investigating how deep learning methods can be made interpretable \cite{kwon2019retainvis} through the proposed UI. For TL, we are investigating whether our solution is generic enough to be applicable to other computer-aided diagnosis problems, especially problems that require multi-class classification. For our proposed loss function, our next step is to investigate how the solution performs for different types of imbalanced classification problems, especially the problems prevalent in computer-aided diagnosis. All three case-studies are accompanied by source code, available at: 1) Case Study I: \url{https://github.com/naimulkhan/SOMVolRen} , 	2) Case Study II: \url{https://github.com/marciahon29/AlzheimersProject/} ,
	3) Case Study III: \url{https://github.com/nabsabraham/focal-tversky-unet} . 

\bibliographystyle{latex8}
\bibliography{MIPR}

\end{document}